\documentclass{article}
\usepackage{spconf, amsmath, amssymb, amsfonts, graphicx, multirow, booktabs, adjustbox, bm, mathtools, url}

\newcommand{\vect}[1]{{\bm {#1}}}

\title{ErpGS: Equirectangular Image Rendering enhanced with 3D Gaussian Regularization}

\name{Shintaro Ito${\dagger}$, Natsuki Takama${\dagger}$, Koichi Ito${\dagger}$, Hwann-Tzong Chen${\dagger\dagger}$, and Takafumi Aoki${\dagger}$}
\address{${\dagger}$Graduate School of Information Sciences, Tohoku University, Japan\\
${\dagger\dagger}$Department of Computer Science, National Tsing Hua University, Taiwan}

\begin{document}
\ninept
\maketitle
\begin{abstract}
  The use of multi-view images acquired by a 360-degree camera can reconstruct a 3D space with a wide area.
  There are 3D reconstruction methods from equirectangular images based on NeRF and 3DGS, as well as Novel View Synthesis (NVS) methods.
  On the other hand, it is necessary to overcome the large distortion caused by the projection model of a 360-degree camera when equirectangular images are used.
  In 3DGS-based methods, the large distortion of the 360-degree camera model generates extremely large 3D Gaussians, resulting in poor rendering accuracy.
  We propose ErpGS, which is Omnidirectional GS based on 3DGS to realize NVS addressing the problems.
  ErpGS introduce some rendering accuracy improvement techniques: geometric regularization, scale regularization, and distortion-aware weights and a mask to suppress the effects of obstacles in equirectangular images.
  Through experiments on public datasets, we demonstrate that ErpGS can render novel view images more accurately than conventional methods.
\end{abstract}

\section{Introduction}
\label{sec:intro}

3D reconstruction from images taken from multiple viewpoints is an essential technique that can be applied to VR/AR, robotics, and 3D map creation, since such images can be captured by a standard camera.
With the rapid development of Novel View Synthesis (NVS) technologies, many methods have been proposed for reconstructing a large space from a large number of images \cite{TancikYPMSBK22,KerblMKWLD24}.
To reduce the time and effort required to acquire a large number of images, the use of a 360-degree camera has attracted attention for its ability to capture the entire circumference of a camera at once and to be easily attached to robots, automobiles, etc.
Since a 360-degree camera images a scene based on Equirectangular Projection (ERP), the image captured by a 360-degree camera is called an ERP image in the following.
So far, there have been several methods proposed for 3D reconstruction from ERP images \cite{HuangCZY22,ChoiKMKM23,LiHYC24,LeeCHM24,HuangBGLG24}.
These methods utilize Neural Radiance Fields (NeRF) \cite{MildenhallSTBRN20} or 3D Gaussian Splatting (GS) \cite{KerblKLD23} to optimize the 3D space representation such as the radiance fields and GS, and then the optimized 3D space representation can be used to render novel view images.
These methods suffer from the problem of low accuracy in NVS since they do not necessarily address the problems inherent in ERP images.

As mentioned above, a standard camera produces images based on perspective projection, while a 360-degree camera produces images based on ERP.
Therefore, ERP images contain strong distortions inherent in a 360-degree camera.
In addition, since the 360-degree camera captures the entire circumference of itself, the robot or stand on which the 360-degree camera is installed appears in the ERP images.
These objects such as robot and stand, appear at different positions in the ERP images depending on the camera position and thus become obstacles that are geometrically inconsistent in the radiance fields or GS.
Therefore, NVS methods based on images acquired with a standard camera cannot be directly applied, and it is necessary to develop an NVS method that addresses such problems inherent in ERP images.

In this paper, we propose Omnidirectional GS based on 3DGS to realize NVS addressing problems inherent in ERP images, which is called {\it ErpGS}.
3D Gaussians obtained from ERP images may contain significantly large Gaussians due to strong distortions in the ERP images.
Since such Gaussians significantly degrade the accuracy of NVS, ErpGS introduces regularization for the scale of the 3D Gaussians.
In the optimization of Omnidirectional GS, it is necessary to take into account the strong distortion in the ERP image.
ErpGS iteratively optimizes Omnidirectional GS so that the normal computed based on the depth map rendered from Omnidirectional GS considering the distortion in the ERP images and the normal directly rendered from Omnidirectional GS become closer.
In addition, ErpGS introduces distortion-aware weights and a mask to suppress the effects of obstacles in the ERP image in the calculation of the loss functions.
Through experiments on public datasets, we demonstrate that ErpGS can render novel view images more accurately than conventional methods.

\begin{figure*}[t]
  \centering
  \includegraphics[width=\linewidth]{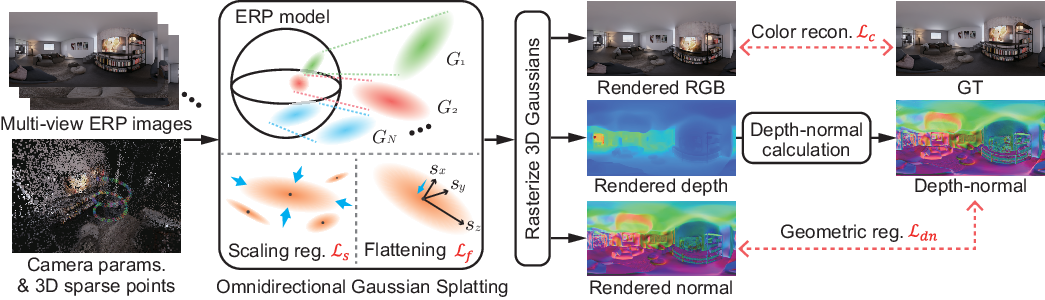}
  \caption{Overview of ErpGS.}
  \label{fig:overview}
\end{figure*}

\section{Related Work}
\label{sec:related_works}

NVS methods such as NeRF \cite{MildenhallSTBRN20} and 3DGS \cite{KerblKLD23} are rapidly developing, which can learn a 3D space representation from multiple images and render photorealistic images.
NVS methods for ERP images based on NeRF have been proposed to render an unknown viewpoint image from a single image \cite{WangWCWLL24,HsuSC21,BaiHGGLG24}.
Although these methods require only a single image and require little effort to acquire data, it is difficult to learn the radiance fields that represent the entire 3D space of the target scene from only a single image.
NVS methods that use multiple viewpoint images \cite{ChoiKMKM23,GeraDRNL22} or videos \cite{ChoiJK24} as input can optimize the radiance fields of a target scene modeled by Multilayer Perceptron (MLP) from multiple viewpoints, and then synthesize novel viewpoint images using optimized MLP.
The NeRF-based methods exhibit high accuracy in rendering novel view images, while training MLP takes several hours.
3DGS \cite{KerblKLD23} can render novel viewpoint images faster and more accurately than NeRF.
There have been several methods proposed to obtain a 3D space representation of a wide area by using Omnidirectional GS supporting the ERP model \cite{LiHYC24,LeeCHM24,HuangBGLG24}.
Due to insufficient support for ERP images, rendering accuracy is degraded by geometric inconsistency among viewpoints and by composing GSs with extremely large 3D Gaussians.
In this paper, we address the above problems of Omnidirectional GS by introducing multiple regularizations that take into account the characteristics of ERP images to improve the rendering accuracy in NVS from ERP images.

\section{ErpGS}
\label{sec:method}

Fig. \ref{fig:overview} shows an overview of ErpGS proposed in this paper.
We describe the main parts of ErpGS as follows: Omnidirectional GS, geometric regularization, scale regularization, and optimization of Omnidirectional GS.

\subsection{Omnidirectional GS}

Omnidirectional GS is used to render images at unknown viewpoints using ERP images taken from multiple viewpoints, associated camera parameters, and a sparse 3D point cloud of the target scene as input \cite{LiHYC24,LeeCHM24,HuangBGLG24}.
Similar to 3DGS \cite{KerblKLD23}, which assumes perspective projection images as input, a novel view image can be synthesized by learning a 3D scene representation from the distribution of 3D Gaussians $\vect{G} = \{ G_i| 1 \leq i \leq N_{g}\}$, where $N_g$ is the number of 3D Gaussians.
Each 3D Gaussian $G_i$ has a 3D position $\vect{\mu}_i$ in the world coordinate system, a scale $\vect{s}_i=(s_x, s_y, s_z)$ for each axis, a quaternion representing rotation $\vect{q}_i$ and opacity $\bm{o}_i$.
The 3D space representation can be learned by iteratively optimizing these parameters of 3D Gaussians.
Using the above parameters and the covariance matrix $\vect{\Sigma}_i$ obtained from $\vect{s}_i$ and $\vect{q}_i$, the 3D Gaussian $G_i$ is given by
\begin{equation}
  G_i(\vect{x}; \vect{\mu}_i, \vect{\Sigma}_i) = e^{-\frac{1}{2}(\vect{x}-\vect{\mu}_i)\vect{\Sigma}^{-1}_i(\vect{x}-\vect{\mu}_i)}.
\end{equation}
The most significant difference from 3DGS is that Omnidirectional GS employs the ERP model as the camera model.
When the center point of 3D Gaussian $G^{c}_i$ in the camera coordinate system, $\vect{\mu}^c_i = (\mu^c_x, \mu^c_y, \mu^c_z)$, is projected onto the image coordinates using ERP model, this transformation is given by
\begin{equation}
  \begin{bmatrix}
    lon \\
    lat
  \end{bmatrix}
  = 
  \begin{bmatrix}
    \arctan 2(\mu^c_x, \mu^c_z) \\
    \arcsin (\mu^c_y, \| \vect{\mu}^c_i \|)
  \end{bmatrix},
\end{equation}
where $lat$ and $lon$ are the coordinates of latitude and longitude, respectively, $-\pi/2 \leq lat \leq \pi/2$, and $-\pi \leq lon < \pi$.
Note that ErpGS assumes that the $x$, $y$, and $z$ axes of the camera coordinate system are oriented right, down, and forward, respectively, and that the $z$ axis is toward the center of the ERP image.
The center point $\mu^p_i$ of $G_i$ in image coordinates is obtained by transforming the coordinates expressed in latitude and longitude to the image coordinate system by
\begin{equation}
  \vect{\mu}^p_i = 
  \begin{bmatrix}
    \frac{W}{2\pi}(lon + \pi) \\
    \frac{H}{2\pi}(2lat + \pi)
  \end{bmatrix},
\end{equation}
where $H$ and $W$ are the height and width of the ERP image, respectively.
The covariance matrix $\vect{\Sigma}^{p}_i$ on the image coordinates is given by the Jacobi matrix $\vect{J}_{erp}$ for the transformation from the camera coordinate system to the image coordinates using the ERP model and the affine approximation \cite{Matthias01} as follows:
\begin{equation}
  \vect{\Sigma}^{p}_i \approx \vect{J}_{erp}\vect{T}\vect{\Sigma}\vect{T}^T\vect{J}^T_{erp},
\end{equation}
where $\vect{T} = [\vect{R}|\vect{t}]$ is a transformation matrix consisting of the rotation matrix $\vect{R}$ from the world coordinate system to the camera coordinate system and the translation vector $\vect{t}$ of a camera position.
$\alpha_i$ is computed as in \cite{LiHYC24} from $\vect{o}_i$ of the 3D Gaussian $G_i$ corresponding to the pixel $\vect{p} \in \mathcal{\vect{P}}$.
Using $\alpha_i$, the RGB value $\mathcal{\vect{C}}(\vect{p})$ of the novel view image can be calculated by
\begin{eqnarray}
  \mathcal{\vect{C}}(\vect{p}) &=& \sum^{N_g}_{i=1}{\vect{c}_i w_i},\\
  w_i &=& \alpha_i \prod^{i-1}_{j=1}{ \{ 1-\alpha_j\} },
\end{eqnarray}
where $w_i$ is the weight of $G_i$ for each pixel and $N_g$ is the total number of 3D Gaussians to be rasterized when rendering the RGB image.

\subsection{Geometric Regularization}
\label{sec:distortion-aware_geometric_regularization}

Conventional Omnidirectional GS \cite{LiHYC24,LeeCHM24,HuangBGLG24} can render photorealistic images, however, it does not guarantee geometric consistency in 3D space.
Therefore, depending on the viewpoint, the rendered RGB image may contain floaters.
Geometric regularization improves rendering accuracy by learning Omnidirectional GS with geometric consistency through regularization using depths and normals.

\noindent
{\bf Normal and Depth Rendering} --- 
Similar to the RGB rendering, the normal of $\vect{p}$ is rendered based on alpha blending as follows:
\begin{equation}
  \mathcal{\vect{N}}(\vect{p}) = \sum^{N_g}_{i=1}{\vect{R} \vect{n}_iw_i},
\end{equation}
where $\vect{n}_i$ is the normal of $G_i$, which is obtained as the unit vector of the smallest eigenvector of $G_i$.
By using the normal, the correct depth to 3D Gaussians can be rendered \cite{ChenLYWXZWLBZ24}.
The depth corresponding $\vect{p}$ is rendered by alpha blending as follows:
\begin{equation}
  \mathcal{D}(\vect{p}) = \frac{1}{\mathcal{\vect{N}}(\vect{p}) \cdot \tilde{\vect{p}}}\sum^{N_g}_{i=1}{d_i w_i},
\end{equation}
where $d_i$ is the distance from the camera center to $G_i$, and $\tilde{\vect{p}}$ is $\vect{p}$ in the homogeneous coordinate system.

\begin{table*}[t]
  \centering
  \caption{Quantitative results of RGB rendering at novel viewpoints compared with EgoNeRF \cite{ChoiKMKM23}, ODGS \cite{LeeCHM24}, OmniGS \cite{LiHYC24} and proposed method (Ours). In this table, LPIPS is based on AlexNet \cite{KrizhevskySH12} to encode rendered images.}
  \label{tab:quantitative_result}
  \begin{adjustbox}{width=\textwidth}
    \begin{tabular}{cccccc|cccc|cccc}
      \toprule
      \multicolumn{2}{c}{} & \multicolumn{4}{c}{PSNR [dB] $\uparrow$} & \multicolumn{4}{c}{SSIM $\uparrow$} & \multicolumn{4}{c}{LPIPS (A) $\downarrow$} \\
      \cmidrule(rl){3-6}
      \cmidrule(rl){7-10}
      \cmidrule(rl){11-14}
      Dataset & 
      \multicolumn{1}{c}{Scene} & \rotatebox{60}{EgoNeRF} & \rotatebox{60}{ODGS} & \rotatebox{60}{OmniGS} & \multicolumn{1}{c}{\rotatebox{60}{Ours}} & \rotatebox{60}{EgoNeRF} & \rotatebox{60}{ODGS} & \rotatebox{60}{OmniGS} & \multicolumn{1}{c}{\rotatebox{60}{Ours}} & \rotatebox{60}{EgoNeRF} & \rotatebox{60}{ODGS} & \rotatebox{60}{OmniGS} & \rotatebox{60}{Ours} \\
      \cmidrule(l){1-1}
      \cmidrule(rl){2-2}
      \cmidrule(l){3-14}
      \multirow{4}{*}{OmniBlender} & 
      barbershop & 30.57 & 33.66 & \underline{37.26} & $\bf 38.71$ & 0.900 & 0.947 & \underline{0.974} & $\bf 0.979$ & 0.187 & 0.123 & \underline{0.050} & $\bf 0.040$ \\
      & lone-monk & \underline{31.10} & 28.58 & 29.00 & $\bf 32.34$ & 0.935 & 0.922 & \underline{0.943} & $\bf 0.963$ & 0.073 & 0.098 & \underline{0.067} & $\bf 0.037$ \\
      & archiviz-flat & 31.69 & 32.50 & \underline{33.38} & $\bf 35.95$ & 0.917 & 0.943 & \underline{0.954} & $\bf 0.963$ & 0.103 & 0.095 & \underline{0.056} & $\bf 0.040$ \\
      & classroom & 26.75 & 26.20 & \underline{33.03} & $\bf 33.62$ & 0.770 & 0.798 & \underline{0.906} & $\bf 0.917$ & 0.368 & 0.385 & \underline{0.190} & $\bf 0.157$ \\
      \cmidrule(l){1-1}
      \cmidrule(rl){2-2}
      \cmidrule(l){3-14}
      \multirow{4}{*}{Ricoh360} & 
      bricks & \underline{24.39} & 22.23 & 22.27 & $\bf 25.03$ & \underline{0.791} & 0.724 & 0.766 & $\bf 0.820$ & \underline{0.186} & 0.293 & 0.251 & $\bf 0.173$ \\
      & center & $\bf 29.42$ & 24.37 & 26.78 & \underline{28.63} & \underline{0.874} & 0.789 & 0.855 & $\bf 0.879$ & \underline{0.144} & 0.396 & 0.188 & $\bf 0.138$ \\
      & farm & $\bf 22.58$ & 20.31 & 20.04 & \underline{21.66} & \underline{0.695} & 0.630 & 0.654 & $\bf 0.696$ & 0.239 & 0.396 & \underline{0.275} & $\bf 0.210$ \\
      & flower & \underline{22.09} & 19.61 & 21.74 & $\bf 22.88$ & 0.658 & 0.597 & 0.715 & $\bf 0.747$ & 0.291 & 0.474 & \underline{0.258} & $\bf 0.208$ \\
      \cmidrule(l){1-1}
      \cmidrule(rl){2-2}
      \cmidrule(l){3-14}
      \multirow{3}{*}{OmniScenes} & 
      pyebaek & 25.05 & 23.82 & \underline{26.67} & $\bf 27.08$ & 0.795 & 0.796 & \underline{0.862} & $\bf 0.867$ & 0.244 & 0.256 & \underline{0.168} & $\bf 0.156$ \\
      & room & 28.69 & 27.25 & \underline{30.29} & $\bf{31.00}$ & 0.904 & 0.900 & \underline{0.928} & $\bf 0.932$ & 0.202 & 0.225 & \underline{0.155} & $\bf 0.142$ \\
      & wedding-hall & 26.00 & 24.94 & \underline{26.99} & $\bf 27.35$ & 0.826 & 0.831 & \underline{0.868} & $\bf 0.872$ & 0.242 & 0.273 & \underline{0.193} & $\bf 0.171$ \\
      \bottomrule
    \end{tabular}
  \end{adjustbox}
\end{table*}
\begin{table}[t]
    \centering
    \caption{Ablation studies for proposed components in terms of the quality of rendered RGB on OmniBlender.}
    \begin{tabular}{lcccc}
        \toprule
          & PSNR $\uparrow$ & \multirow{2}{*}{SSIM $\uparrow$} & \multirow{2}{*}{LPIPS (A) $\downarrow$} & \multirow{2}{*}{LPIPS (V) $\downarrow$} \\
         Ablation & [dB] &  &  &   \\
         \cmidrule(lr){1-1}
         \cmidrule(lr){2-5}
          w/o $\mathcal{\vect{W}}$ & 33.86 & 0.949 & 0.0852 & 0.1630 \\
          w/o $\mathcal{L}_{dn}$ & 34.61 & 0.953 & 0.0717 & 0.1429 \\
          w/o $\mathcal{L}_{s}$ & 34.64 & 0.954 & 0.0723 &  0.1434\\
          All & $\bf 35.16$ & $\bf 0.956$ & $\bf 0.0687$ & $\bf 0.1374$ \\
         \bottomrule
    \end{tabular}
    \label{tab:ablation_study}
\end{table}

\begin{figure*}[t]
  \centering
  \includegraphics[width=.95\linewidth]{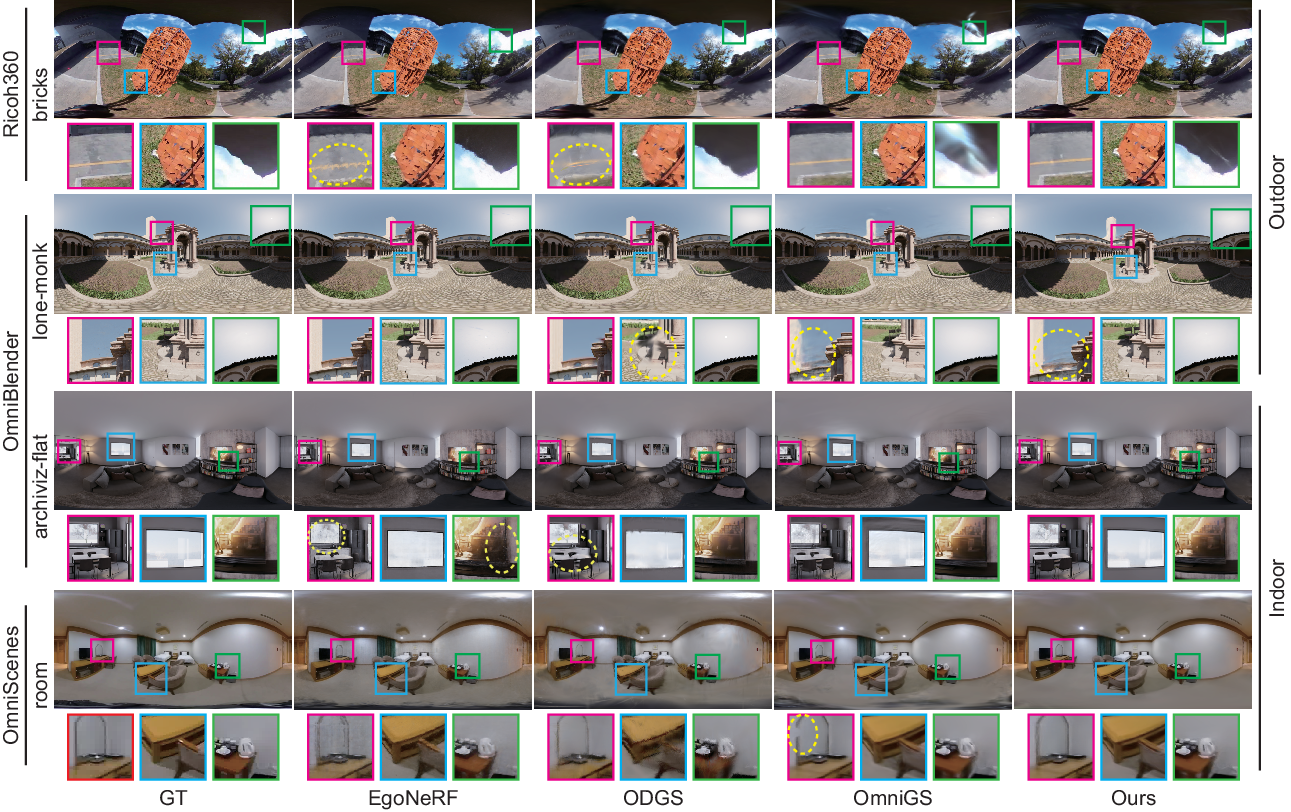}
  \caption{Experimental results of rendered ERP images at novel viewpoints on several datasets.}
  \label{fig:main-qualitative-results}
\end{figure*}
\begin{figure*}[t!]
  \centering
  \includegraphics[width=.95\linewidth]{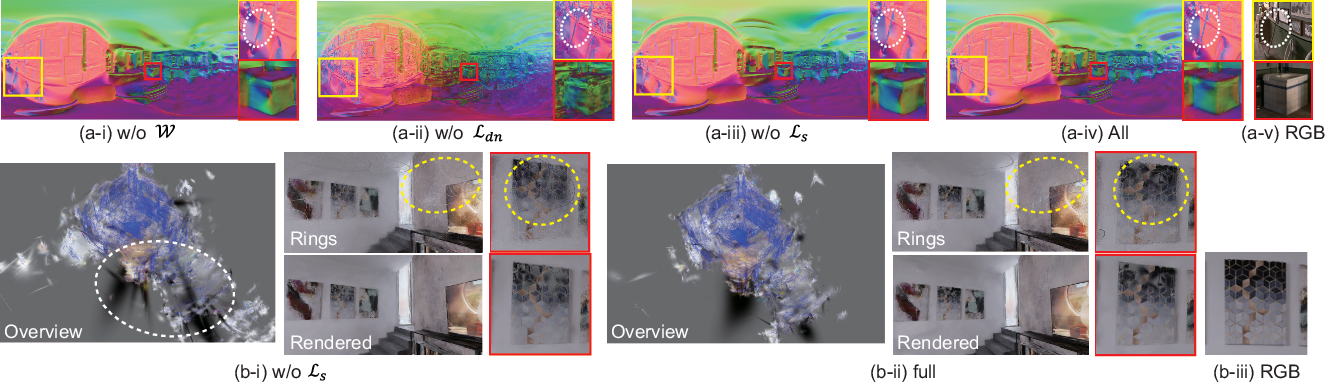}
  \caption{Qualitative results of ablation studies: (a) Rendered normal maps, (b) Gaussian ellipsoids and rendered RGB. `Overview' means 3D Gaussians seen from a distance in a target scene. Blue points depict centers of each 3D Gaussian. `Rings' means the visualized 3D Gaussians with rings. In the result of w/o $\mathcal{L}_s$, rendered quality degraded due to large 3D Gaussians.}
  \label{fig:ablation-study_qualitative}
\end{figure*}

\noindent
{\bf Omnidirectional Neighbor Pixel Selection} --- 
The normal can be calculated from the depth by projecting a point on the image plane into 3D space using the depth corresponding to the pixel of interest and its neighbors \cite{ChenLYWXZWLBZ24}.
Although the pixels neighboring the pixel of interest can be considered as neighbors in the perspective projection image, the pixels neighboring the pixel of interest are not necessarily correct neighbors on the ERP image \cite{CoorsCG18}.
On the unit sphere, which is the image plane of the ERP model, the pixel neighboring the pixel of interest can be selected from the neighboring pixels on the tangent plane centered on the pixel of interest.
When the pixel of interest is located at $(0, 0)$ on the tangent plane, the correspondence between the adjacent pixels on the tangent plane and the pixels in the ERP image can be represented as $\vect{t}_{(\pm 1, 0)} = (\pm \tan(2 \pi / W), 0)$, $\vect{t}_{(0, \pm 1)} = (0, \pm \tan(2 \pi / H))$.
Using these $x$ components, $t_x$, and $y$ components, $t_y$, adjacent pixels in the ERP image can be selected as
\begin{eqnarray}
  \Phi(t_x, t_y) &=& \arcsin \left( \cos\nu \sin\phi + 
                          \frac{t_y\sin\nu \cos\phi}{\rho} \right), \\
  \Theta(t_x, t_y) &=& \theta + \arctan \left( \frac{t_x \sin\nu}{\rho\cos\phi \cos\nu - t_y\sin\phi \sin\nu} \right),
\end{eqnarray}
where $\nu = \arctan(\rho)$ and $\rho=\sqrt{{t_x}^2 + {t_y}^2}$.

\noindent
{\bf Depth-normal calculation} --- 
Using the image coordinates of the pixel of interest and the adjacent pixel and the depth corresponding to each pixel, the 3D positions $\vect{P}_{(\pm1,0)}, \vect{P}_{(0, \pm1)}$ corresponding to the adjacent pixel $(\pm1, 0), (0, \pm1)$ in the tangent plane can be calculated.
The normal vector in the camera coordinate system is obtained by
\begin{equation}
  \mathcal{\vect{N}}_d(\vect{p}) = \frac{(\vect{P}_{(1,0)} - \vect{P}_{(-1,0)}) \times (\vect{P}_{(0,1)} - \vect{P}_{(0,-1)})}{|(\vect{P}_{(1,0)} - \vect{P}_{(-1,0)}) \times (\vect{P}_{(0,1)} - \vect{P}_{(0,-1)})|}.
\end{equation}
The error between the depth-normal $\mathcal{\vect{N}}_{d}(\vect{p})$ calculated from the depth and the normal $\mathcal{\vect{N}}(\vect{p})$ directly rendered from 3D Gaussians is calculated by
\begin{equation}
  DNE(\vect{p}) = |\overline{\nabla \mathcal{I}(\vect{p})}|^2 \| \mathcal{\vect{N}}_d(\vect{p}) - \mathcal{N}(\vect{p}) \|,
\end{equation}
where $| \overline{\nabla \mathcal{I}(\vect{p})} |$ is the color gradient.
The direction of the gradient calculation is also determined by selecting neighboring pixels in the ERP image using the method described above.

\subsection{Scale Regularization}

NVS methods that assume ERP images as input, such as OmniGS \cite{LiHYC24} and ODGS \cite{LeeCHM24}, can render novel view images both fast and accurately.
On the other hand, we found that extremely large 3D Gaussians are generated to handle the distortions inherent in ERP images, resulting in 3D inconsistency in the distribution of the 3D Gaussians and negatively affecting the optimization of the 3D Gaussians and the final rendering accuracy.
OmniGS \cite{LiHYC24} avoids this problem by setting a threshold to the size of 3D Gaussians for each scene and pruning out Gaussians that are larger than the threshold.
ErpGS improves rendering accuracy by introducing a regularization term for the scale of 3D Gaussians into the loss, while suppressing the generation of 3D Gaussians with too large a size for the scene.
In addition, to make it easier to estimate the normals at each viewpoint, ErpGS introduces a loss to flatten the 3D Gaussians, as in \cite{ChenLL23,ChenLYWXZWLBZ24}, which is given by
\begin{equation}
  \mathcal{L}_{s} = \frac{1}{N_{g}}\sum^{N_{g}}_{i=1} \| \vect{s}_i \|^{2}_{2}, \quad \mathcal{L}_{f} = \| \min (s_x, s_y, s_z) \|_{1},
\end{equation}

\subsection{Optimization}

The following describes the techniques used in the optimization of Omnidirectional GS in ErpGS.

\noindent
{\bf Distortion-aware Weight} ---
The higher the latitude, the greater the ERP distortion in the ERP image.
Therefore, at high latitudes, the area in 3D space for each pixel in the ERP image becomes smaller, while at low latitudes it becomes larger \cite{OtonariIA22}.
We add weight $\mathcal{\bm{W}}$ to the rendered RGB image, depth map, and normal map, taking into account the distortion of the ERP image, where $\mathcal{\bm{W}}$ is given by
\begin{equation}
  \mathcal{\bm W} = \int^{\theta_{1}}_{\theta_{0}}\int^{\phi_{1}}_{\phi_{0}}{cos\theta d\theta d\phi}
\end{equation}

\noindent
{\bf Viewpoint-dependent Mask} --- 
When capturing images with a 360-degree camera, the photographer, the robot on which the 360-degree camera is mounted, and the camera stick or stand appear at different positions from different viewpoints.
These effects can interfere with the consistency of the 3D scene when learning Omnidirectional GS.
ErpGS introduces a viewpoint-dependent mask $\mathcal{\vect{M}}_\vect{p}$ to loss functions instead of a viewpoint-independent mask to reduce the effect of obstacles that interfere with Gaussian optimization.
The loss functions with $\mathcal{\vect{M}}_\vect{p}$ are given by
\begin{equation}
  \mathcal{L}_{color} = \frac{\sum_{\vect{p} \in \mathcal{\vect{P}}} \{ \mathcal{\vect{W}}_{\vect{p}} \odot \mathcal{\vect{M}}_{\vect{p}} \odot CRE(\vect{p}) \}}{\sum_{\vect{p} \in \mathcal{\vect{P}}} \{\mathcal{\vect{W}}_{\vect{p}} \odot \mathcal{\vect{M}}_{\vect{p}}\} },
\end{equation}
\begin{equation}
  \mathcal{L}_{dn} = \frac{\sum_{\vect{p} \in \mathcal{\vect{P}}}\{ \mathcal{\vect{W}}_{\vect{p}} \odot \mathcal{\vect{M}}_{\vect{p}} \odot DNE(\vect{p}) \}}{\sum_{\vect{p} \in \mathcal{\vect{P}}} \{\mathcal{\vect{W}}_{\vect{p}} \odot \mathcal{\vect{M}}_{\vect{p}}\} },
\end{equation}
\begin{equation}
    \begin{split}
        CRE(\vect{p}) &= (1 -\lambda)\| \mathcal{C}(\vect{p}) - \mathcal{C}_{gt}(\vect{p}) \|_{1} \\
        & \quad + \lambda \Big(1-SSIM \big(\mathcal{C}(\vect{p}), \mathcal{C}_{gt}(\vect{p}) \big) \Big),
    \end{split}
\end{equation}
where $\mathcal{C}_{gt}(\vect{p})$ is the ground-truth RGB value of $\vect{p}$ and $SSIM(\cdot, \cdot)$ is a function to calculate Structural Similarity.

\noindent
{\bf Loss Function} ---
The total loss function for ErpGS is given by
\begin{equation}
  \mathcal{L} = \mathcal{L}_{c} + \lambda_{dn} \mathcal{L}_{dn} + \lambda_{f} \mathcal{L}_{f} + \frac{1}{2}\lambda_{s} \mathcal{L}_{s},
\end{equation}
where $\lambda_{dn}$, $\lambda_{f}$, and $\lambda_{s}$ are weights.

\section{Experiments}
\label{sec:experiments}

We demonstrate the effectiveness of ErpGS for NVS using public datasets.

\subsection{Experimental Setup}

\noindent
{\bf Dataset} ---
In the experiments, we use the three public datasets: OmniBlender \cite{ChoiKMKM23}, Ricoh 360 \cite{ChoiKMKM23}, and OmniScenes \cite{KimCJK21}.
OmniBlender \cite{ChoiKMKM23} consists of indoor/outdoor scenes synthesized from Blender projects \cite{Blender}.
Ricoh 360 \cite{ChoiKMKM23} consists of outdoor scenes captured in the real world.
OmniBlender and Ricoh 360 contain egocentric images taken by moving the camera in a spiral motion \cite{ChoiKMKM23}.
OmniScenes \cite{KimCJK21} consists of real-world indoor scenes taken from various positions, i.e., non-egocentric images.
Experiments are conducted using the above datasets with preprocessing by the authors of ODGS \cite{LeeCHM24}.

\noindent
{\bf Baselines} ---
In this experiment, we compare the proposed method, ErpGS, with OmniGS \cite{LiHYC24}, ODGS \cite{LeeCHM24}, and EgoNeRF \cite{ChoiKMKM23}.
OmniGS and ODGS are NVS methods for ERP images based on 3DGS.
Because the code for OmniGS was not publicly available at the time of writing, the OmniGS used in this experiment was reproduced and implemented by the authors.
The 3DGS-based methods, ErpGS, OmniGS, and ODGS, set the number of optimization iterations to 30,000.
EgoNeRF is a NeRF-based method that uses two spherical feature grids and an environment map to efficiently estimate the radiance fields in an unbounded scene.
EgoNeRF trains the model by iteratively optimizing the radiance fields 200,000 times.
All experiments are conducted on NVIDIA GeForce RTX 4090 GPUs (24GB).

\noindent
{\bf Implementation Details} ---
ErpGS is implemented using PyTorch based on the public implementation of 3DGS \cite{KerblKLD23}.
The rasterizer for Omnidirectional GS is implemented in CUDA for faster speed.
Adam \cite{KingmaB14} is used as an optimizer.
The hyperparameters used for optimization are based on the same parameters as for 3DGS \cite{KerblKLD23}.
$\mathcal{L}_{s}$ is introduced from the beginning of the optimization.
After 10,000 optimization iterations $\mathcal{L}_{dn}$ and $\mathcal{L}_{f}$ are added.
We set $\lambda_{dn} = 0.01$, $\lambda_{f} = 100$, and $\lambda_{s} = 0.01$.
Only in the experiments on OmniScenes, we introduce a mask for ErpGS.
When the mask is introduced, the accuracy is evaluated only in the unmasked region for all methods.

\subsection{ERP Image Rendering}

\noindent
{\bf Quantitative Results} ---
In this experiment, PSNR, SSIM, and LPIPS \cite{lpips}, which measure the similarity between the rendered image and the ground-truth image, are used as evaluation metrics.
LPIPS is denoted by LPIPS (A) when AlexNet \cite{KrizhevskySH12} is used as the feature extractor and LPIPS (V) when VGG \cite{LiuD15} is used.
Table \ref{tab:quantitative_result} shows the quantitative results of NVS for each method.
ErpGS exhibits better rendering performance than the other methods on all datasets.
While EgoNeRF has high rendering accuracy for ego-centric data taken in outdoors, such as Ricoh 360, ErpGS also renders RGB images with equal or better accuracy.
ErpGS exhibits high rendering accuracy even for non-egocentric images such as OmniScenes.

\noindent
{\bf Qualitative Results} ---
Fig. \ref{fig:main-qualitative-results} shows the novel view images rendered by each method.
ErpGS achieves better rendering accuracy than other methods for both outdoor and indoor scenes.
The results of `lone-monk' in OmniBlender show that EgoNeRF can learn robustly even in low-texture regions such as the sky, since it creates an environment map in the learning process, and can render the tower in the back clearly.
On the other hand, ErpGS cannot render the area near the tower.
This reason may be due to the small number of initial 3D point clouds that exist near the tower.
EgoNeRF, ODGS, and OmniGS are strongly affected by the effect of the stand, as in the case of a `room', resulting in low rendering accuracy.
By applying masks to such regions, ErpGS can suppress negative influences from objects that are unnecessary for 3D representation and maintain high rendering accuracy for novel views.

\subsection{Ablation Study}
Table \ref{tab:ablation_study} shows the results of the ablation study in OmniBlender.
Removing each element in ErpGS reduces the accuracy of NVS.
In particular, removing the weight $\mathcal{\vect{W}}$ for each pixel corresponding to the distortion of the ERP images reduces the rendering accuracy, and therefore it is important to optimize considering the characteristics of the ERP images.
The quantitative accuracy of the rendering is only slightly improved by introducing a regularization term.
As shown in Fig.\ref{fig:ablation-study_qualitative} (a-i$\sim$iv), the rendered normals are not smooth when $\mathcal{L}_{dn}$ and $\mathcal{L}_{s}$ are excluded.
Therefore, the introduction of the regularization proposed in this paper can significantly improve the rendering accuracy of normal maps.
Also, as shown in Fig. \ref{fig:ablation-study_qualitative} (b-i), removing the regularization term from the proposed method removes large 3D Gaussians and improves the rendering accuracy.

\section{Conclusion}
\label{sec:conclusion}

We proposed {\it ErpGS}, which is Omnidirectional GS based on 3DGS to realize NVS addressing problems inherent in ERP images.
ErpGS introduced some rendering accuracy improvement techniques: geometric regularization, scale regularization, and distortion-aware weights and a mask to suppress the effects of obstacles in the ERP images.
Through experiments on public datasets, we demonstrated that ErpGS can render novel view images more accurately than conventional methods.

\section{Acknowledgment}
This work was supported in part by JSPS KAKENHI 23H00463 and 25K03131, and JST BOOST JPMJBS2421.

{\small
  \bibliographystyle{IEEEbib}
  \bibliography{strings}
}
\end{document}